# SMOTE: Synthetic Minority Over-sampling Technique

**Nitesh V. Chawla**                         CHAWLA@CSEE.USF.EDU
*Department of Computer Science and Engineering, ENB 118*
*University of South Florida*
*4202 E. Fowler Ave.*
*Tampa, FL 33620-5399, USA*

**Kevin W. Bowyer**                         KWB@CSE.ND.EDU
*Department of Computer Science and Engineering*
*384 Fitzpatrick Hall*
*University of Notre Dame*
*Notre Dame, IN 46556, USA*

**Lawrence O. Hall**                         HALL@CSEE.USF.EDU
*Department of Computer Science and Engineering, ENB 118*
*University of South Florida*
*4202 E. Fowler Ave.*
*Tampa, FL 33620-5399, USA*

**W. Philip Kegelmeyer**                  WPK@CALIFORNIA.SANDIA.GOV
*Sandia National Laboratories*
*Biosystems Research Department, P.O. Box 969, MS 9951*
*Livermore, CA, 94551-0969, USA*

## Abstract

An approach to the construction of classifiers from imbalanced datasets is described. A dataset is imbalanced if the classification categories are not approximately equally represented. Often real-world data sets are predominately composed of "normal" examples with only a small percentage of "abnormal" or "interesting" examples. It is also the case that the cost of misclassifying an abnormal (interesting) example as a normal example is often much higher than the cost of the reverse error. Under-sampling of the majority (normal) class has been proposed as a good means of increasing the sensitivity of a classifier to the minority class. This paper shows that a combination of our method of over-sampling the minority (abnormal) class and under-sampling the majority (normal) class can achieve better classifier performance (in ROC space) than only under-sampling the majority class. This paper also shows that a combination of our method of over-sampling the minority class and under-sampling the majority class can achieve better classifier performance (in ROC space) than varying the loss ratios in Ripper or class priors in Naive Bayes. Our method of over-sampling the minority class involves creating synthetic minority class examples. Experiments are performed using C4.5, Ripper and a Naive Bayes classifier. The method is evaluated using the area under the Receiver Operating Characteristic curve (AUC) and the ROC convex hull strategy.

## 1. Introduction

A dataset is imbalanced if the classes are not approximately equally represented. Imbalance on the order of 100 to 1 is prevalent in fraud detection and imbalance of up to 100,000 to 1 has been reported in other applications (Provost & Fawcett, 2001). There have been attempts to deal with imbalanced datasets in domains such as fraudulent telephone calls (Fawcett & Provost, 1996), telecommunications management (Ezawa, Singh, & Norton, 1996), text classification (Lewis & Catlett, 1994; Dumais, Platt, Heckerman, & Sahami, 1998; Mladenić & Grobelnik, 1999; Lewis & Ringuette, 1994; Cohen, 1995a) and detection of oil spills in satellite images (Kubat, Holte, & Matwin, 1998).

The performance of machine learning algorithms is typically evaluated using predictive accuracy. However, this is not appropriate when the data is imbalanced and/or the costs of different errors vary markedly. As an example, consider the classification of pixels in mammogram images as possibly cancerous (Woods, Doss, Bowyer, Solka, Priebe, & Kegelmeyer, 1993). A typical mammography dataset might contain 98% normal pixels and 2% abnormal pixels. A simple default strategy of guessing the majority class would give a predictive accuracy of 98%. However, the nature of the application requires a fairly high rate of correct detection in the minority class and allows for a small error rate in the majority class in order to achieve this. Simple predictive accuracy is clearly not appropriate in such situations. The Receiver Operating Characteristic (ROC) curve is a standard technique for summarizing classifier performance over a range of tradeoffs between true positive and false positive error rates (Swets, 1988). The Area Under the Curve (AUC) is an accepted traditional performance metric for a ROC curve (Duda, Hart, & Stork, 2001; Bradley, 1997; Lee, 2000). The ROC convex hull can also be used as a robust method of identifying potentially optimal classifiers (Provost & Fawcett, 2001). If a line passes through a point on the convex hull, then there is no other line with the same slope passing through another point with a larger true positive (TP) intercept. Thus, the classifier at that point is optimal under any distribution assumptions in tandem with that slope.

The machine learning community has addressed the issue of class imbalance in two ways. One is to assign distinct costs to training examples (Pazzani, Merz, Murphy, Ali, Hume, & Brunk, 1994; Domingos, 1999). The other is to re-sample the original dataset, either by over-sampling the minority class and/or under-sampling the majority class (Kubat & Matwin, 1997; Japkowicz, 2000; Lewis & Catlett, 1994; Ling & Li, 1998). Our approach (Chawla, Bowyer, Hall, & Kegelmeyer, 2000) blends under-sampling of the majority class with a special form of over-sampling the minority class. Experiments with various datasets and the C4.5 decision tree classifier (Quinlan, 1992), Ripper (Cohen, 1995b), and a Naive Bayes Classifier show that our approach improves over other previous re-sampling, modifying loss ratio, and class priors approaches, using either the AUC or ROC convex hull.

Section 2 gives an overview of performance measures. Section 3 reviews the most closely related work dealing with imbalanced datasets. Section 4 presents the details of our approach. Section 5 presents experimental results comparing our approach to other re-sampling approaches. Section 6 discusses the results and suggests directions for future work.

## 2. Performance Measures

The performance of machine learning algorithms is typically evaluated by a confusion matrix as illustrated in Figure 1 (for a 2 class problem). The columns are the Predicted class and the rows are the Actual class. In the confusion matrix, $TN$ is the number of negative examples correctly classified (True Negatives), $FP$ is the number of negative examples incorrectly classified as positive (False Positives), $FN$ is the number of positive examples incorrectly classified as negative (False Negatives) and $TP$ is the number of positive examples correctly classified (True Positives).

|  | Predicted Negative | Predicted Positive |
|---|---|---|
| Actual Negative | TN | FP |
| Actual Positive | FN | TP |

Figure 1: Confusion Matrix

Predictive accuracy is the performance measure generally associated with machine learning algorithms and is defined as $Accuracy = (TP + TN)/(TP + FP + TN + FN)$. In the context of balanced datasets and equal error costs, it is reasonable to use error rate as a performance metric. Error rate is $1 - Accuracy$. In the presence of imbalanced datasets with unequal error costs, it is more appropriate to use the ROC curve or other similar techniques (Ling & Li, 1998; Drummond & Holte, 2000; Provost & Fawcett, 2001; Bradley, 1997; Turney, 1996).

ROC curves can be thought of as representing the family of best decision boundaries for relative costs of TP and FP. On an ROC curve the X-axis represents $\%FP = FP/(TN+FP)$ and the Y-axis represents $\%TP = TP/(TP+FN)$. The ideal point on the ROC curve would be (0,100), that is all positive examples are classified correctly and no negative examples are misclassified as positive. One way an ROC curve can be swept out is by manipulating the balance of training samples for each class in the training set. Figure 2 shows an illustration. The line y = x represents the scenario of randomly guessing the class. Area Under the ROC Curve (AUC) is a useful metric for classifier performance as it is independent of the decision criterion selected and prior probabilities. The AUC comparison can establish a dominance relationship between classifiers. If the ROC curves are intersecting, the total AUC is an average comparison between models (Lee, 2000). However, for some specific cost and class distributions, the classifier having maximum AUC may in fact be suboptimal. Hence, we also compute the ROC convex hulls, since the points lying on the ROC convex hull are potentially optimal (Provost, Fawcett, & Kohavi, 1998; Provost & Fawcett, 2001).

## 3. Previous Work: Imbalanced datasets

Kubat and Matwin (1997) selectively under-sampled the majority class while keeping the original population of the minority class. They have used the geometric mean as a performance measure for the classifier, which can be related to a single point on the ROC curve. The minority examples were divided into four categories: some noise overlapping the positive class decision region, borderline samples, redundant samples and safe samples. The borderline examples were detected using the Tomek links concept (Tomek, 1976). Another related work proposed the SHRINK system that classifies an overlapping region of minority (positive) and majority (negative) classes as positive; it searches for the "best positive region" (Kubat et al., 1998).

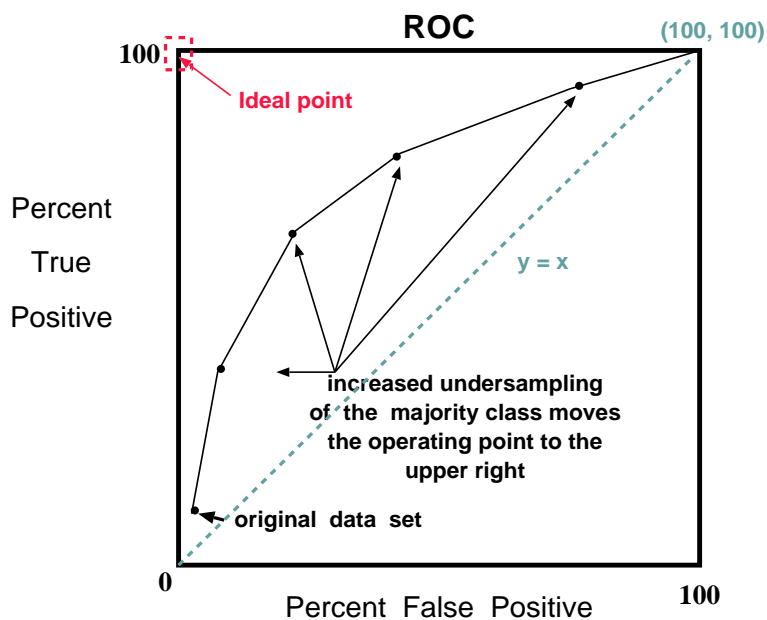

Figure 2: Illustration of sweeping out a ROC curve through under-sampling. Increased under-sampling of the majority (negative) class will move the performance from the lower left point to the upper right.

Japkowicz (2000) discussed the effect of imbalance in a dataset. She evaluated three strategies: under-sampling, resampling and a recognition-based induction scheme. We focus on her sampling approaches. She experimented on artificial 1D data in order to easily measure and construct concept complexity. Two resampling methods were considered. Random resampling consisted of resampling the smaller class at random until it consisted of as many samples as the majority class and "focused resampling" consisted of resampling only those minority examples that occurred on the boundary between the minority and majority classes. Random under-sampling was considered, which involved under-sampling the majority class samples at random until their numbers matched the number of minority class samples; focused under-sampling involved under-sampling the majority class samples lying further away. She noted that both the sampling approaches were effective, and she also observed that using the sophisticated sampling techniques did not give any clear advantage in the domain considered (Japkowicz, 2000).

One approach that is particularly relevant to our work is that of Ling and Li (1998). They combined over-sampling of the minority class with under-sampling of the majority class. They used lift analysis instead of accuracy to measure a classifier's performance. They proposed that the test examples be ranked by a confidence measure and then lift be used as the evaluation criteria. A lift curve is similar to an ROC curve, but is more tailored for the marketing analysis problem (Ling & Li, 1998). In one experiment, they under-sampled the majority class and noted that the best lift index is obtained when the classes are equally represented (Ling & Li, 1998). In another experiment, they over-sampled the positive (minority) examples with replacement to match the number of negative (majority) examples to the number of positive examples. The over-sampling and under-sampling combination did not provide significant improvement in the lift index. However, our approach to over-sampling differs from theirs.

Solberg and Solberg (1996) considered the problem of imbalanced data sets in oil slick classification from SAR imagery. They used over-sampling and under-sampling techniques to improve the classification of oil slicks. Their training data had a distribution of 42 oil slicks and 2,471 look-alikes, giving a prior probability of 0.98 for look-alikes. This imbalance would lead the learner (without any appropriate loss functions or a methodology to modify priors) to classify almost all look-alikes correctly at the expense of misclassifying many of the oil slick samples (Solberg & Solberg, 1996). To overcome this imbalance problem, they over-sampled (with replacement) 100 samples from the oil slick, and they randomly sampled 100 samples from the non oil slick class to create a new dataset with equal probabilities. They learned a classifier tree on this balanced data set and achieved a 14% error rate on the oil slicks in a leave-one-out method for error estimation; on the look alikes they achieved an error rate of 4% (Solberg & Solberg, 1996).

Another approach that is similar to our work is that of Domingos (1999). He compares the "metacost" approach to each of majority under-sampling and minority over-sampling. He finds that metacost improves over either, and that under-sampling is preferable to minority over-sampling. Error-based classifiers are made cost-sensitive. The probability of each class for each example is estimated, and the examples are relabeled optimally with respect to the misclassification costs. The relabeling of the examples expands the decision space as it creates new samples from which the classifier may learn (Domingos, 1999).

A feed-forward neural network trained on an imbalanced dataset may not learn to discriminate enough between classes (DeRouin, Brown, Fausett, & Schneider, 1991). The authors proposed that the learning rate of the neural network be adapted to the statistics of class representation in the data. They calculated an attention factor from the proportion of samples presented to the neural network for training. The learning rate of the network elements was adjusted based on the attention factor. They experimented on an artificially generated training set and on a real-world training set, both with multiple (more than two) classes. They compared this to the approach of replicating the minority class samples to balance the data set used for training. The classification accuracy on the minority class was improved.

Lewis and Catlett (1994) examined heterogeneous uncertainty sampling for supervised learning. This method is useful for training samples with uncertain classes. The training samples are labeled incrementally in two phases and the uncertain instances are passed on to the next phase. They modified C4.5 to include a loss ratio for determining the class values at the leaves. The class values were determined by comparison with a probability threshold of $LR/(LR + 1)$, where $LR$ is the loss ratio (Lewis & Catlett, 1994).

The information retrieval (IR) domain (Dumais et al., 1998; Mladenić & Grobelnik, 1999; Lewis & Ringuette, 1994; Cohen, 1995a) also faces the problem of class imbalance in the dataset. A document or web page is converted into a bag-of-words representation;

that is, a feature vector reflecting occurrences of words in the page is constructed. Usually, there are very few instances of the interesting category in text categorization. This over-representation of the negative class in information retrieval problems can cause problems in evaluating classifiers' performances. Since error rate is not a good metric for skewed datasets, the classification performance of algorithms in information retrieval is usually measured by *precision* and *recall*:

$$recall = \frac{TP}{TP + FN}$$

$$precision = \frac{TP}{TP + FP}$$

Mladenić and Grobelnik (1999) proposed a feature subset selection approach to deal with imbalanced class distribution in the IR domain. They experimented with various feature selection methods, and found that the *odds ratio* (van Rijsbergen, Harper, & Porter, 1981) when combined with a Naive Bayes classifier performs best in their domain. *Odds ratio* is a probabilistic measure used to rank documents according to their relevance to the positive class (minority class). *Information gain* for a word, on the other hand, does not pay attention to a particular target class; it is computed per word for each class. In an imbalanced text dataset (assuming 98 to 99% is the negative class), most of the features will be associated with the negative class. *Odds ratio* incorporates the target class information in its metric giving better results when compared to *information gain* for text categorization.

Provost and Fawcett (1997) introduced the ROC convex hull method to estimate the classifier performance for imbalanced datasets. They note that the problems of unequal class distribution and unequal error costs are related and that little work has been done to address either problem (Provost & Fawcett, 2001). In the ROC convex hull method, the ROC space is used to separate classification performance from the class and cost distribution information.

To summarize the literature, under-sampling the majority class enables better classifiers to be built than over-sampling the minority class. A combination of the two as done in previous work does not lead to classifiers that outperform those built utilizing only under-sampling. However, the over-sampling of the minority class has been done by sampling with replacement from the original data. Our approach uses a different method of over-sampling.

## 4. SMOTE: Synthetic Minority Over-sampling TEchnique

### 4.1 Minority over-sampling with replacement

Previous research (Ling & Li, 1998; Japkowicz, 2000) has discussed over-sampling with replacement and has noted that it doesn't significantly improve minority class recognition. We interpret the underlying effect in terms of decision regions in feature space. Essentially, as the minority class is over-sampled by increasing amounts, the effect is to identify similar but more specific regions in the feature space as the decision region for the minority class. This effect for decision trees can be understood from the plots in Figure 3.

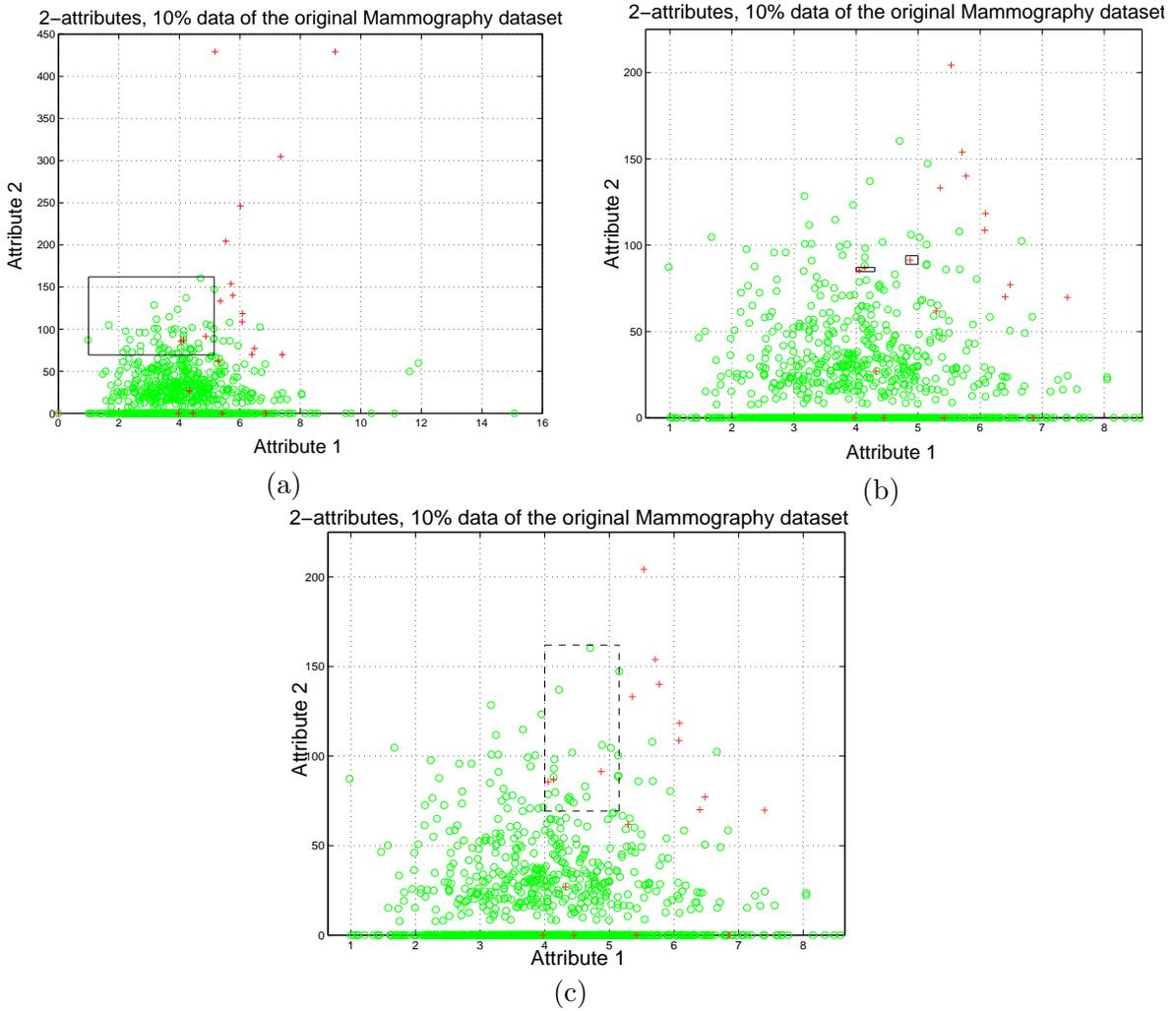

Figure 3: a) Decision region in which the three minority class samples (shown by '+') reside after building a decision tree. This decision region is indicated by the solid-line rectangle. b) A zoomed-in view of the chosen minority class samples for the same dataset. Small solid-line rectangles show the decision regions as a result of over-sampling the minority class with replication. c) A zoomed-in view of the chosen minority class samples for the same dataset. Dashed lines show the decision region after over-sampling the minority class with synthetic generation.

The data for the plot in Figure 3 was extracted from a Mammography dataset[1] (Woods et al., 1993). The minority class samples are shown by + and the majority class samples are shown by $o$ in the plot. In Figure 3(a), the region indicated by the solid-line rectangle is a majority class decision region. Nevertheless, it contains three minority class samples shown by '+' as false negatives. If we replicate the minority class, the decision region for the minority class becomes very specific and will cause new splits in the decision tree. This will lead to more terminal nodes (leaves) as the learning algorithm tries to learn more and more specific regions of the minority class; in essence, overfitting. Replication of the minority class does not cause its decision boundary to spread into the majority class region. Thus, in Figure 3(b), the three samples previously in the majority class decision region now have very specific decision regions.

## 4.2 SMOTE

We propose an over-sampling approach in which the minority class is over-sampled by creating "synthetic" examples rather than by over-sampling with replacement. This approach is inspired by a technique that proved successful in handwritten character recognition (Ha & Bunke, 1997). They created extra training data by performing certain operations on real data. In their case, operations like rotation and skew were natural ways to perturb the training data. We generate synthetic examples in a less application-specific manner, by operating in "feature space" rather than "data space". The minority class is over-sampled by taking each minority class sample and introducing synthetic examples along the line segments joining any/all of the $k$ minority class nearest neighbors. Depending upon the amount of over-sampling required, neighbors from the $k$ nearest neighbors are randomly chosen. Our implementation currently uses five nearest neighbors. For instance, if the amount of over-sampling needed is 200%, only two neighbors from the five nearest neighbors are chosen and one sample is generated in the direction of each. Synthetic samples are generated in the following way: Take the difference between the feature vector (sample) under consideration and its nearest neighbor. Multiply this difference by a random number between 0 and 1, and add it to the feature vector under consideration. This causes the selection of a random point along the line segment between two specific features. This approach effectively forces the decision region of the minority class to become more general.

Algorithm *SMOTE*, on the next page, is the pseudo-code for SMOTE. Table 4.2 shows an example of calculation of random synthetic samples. The amount of over-sampling is a parameter of the system, and a series of ROC curves can be generated for different populations and ROC analysis performed.

The synthetic examples cause the classifier to create larger and less specific decision regions as shown by the dashed lines in Figure 3(c), rather than smaller and more specific regions. More general regions are now learned for the minority class samples rather than those being subsumed by the majority class samples around them. The effect is that decision trees generalize better. Figures 4 and 5 compare the minority over-sampling with replacement and SMOTE. The experiments were conducted on the mammography dataset. There were 10923 examples in the majority class and 260 examples in the minority class originally. We have approximately 9831 examples in the majority class and 233 examples in the minority class for the training set used in 10-fold cross-validation. The minority class was over-sampled at 100%, 200%, 300%, 400% and 500% of its original size. The graphs show that the tree sizes for minority over-sampling with replacement at higher degrees of replication are much greater than those for SMOTE, and the minority class recognition of the minority over-sampling with replacement technique at higher degrees of replication isn't as good as SMOTE.

---

1. The data is available from the USF Intelligent Systems Lab, http://morden.csee.usf.edu/~chawla.

**Algorithm** *SMOTE*(T, N, k)

**Input:** Number of minority class samples $T$; Amount of SMOTE $N\%$; Number of nearest
    neighbors $k$

**Output:** $(N/100)$ * $T$ synthetic minority class samples

1.  (* *If N is less than 100%, randomize the minority class samples as only a random
    percent of them will be SMOTEd.* *)
2.  **if** $N < 100$
3.    **then** Randomize the $T$ minority class samples
4.        $T = (N/100) * T$
5.        $N = 100$
6.  **endif**
7.  $N = (int)(N/100)$ (* *The amount of SMOTE is assumed to be in integral multiples of
    100.* *)
8.  $k$ = Number of nearest neighbors
9.  *numattrs* = Number of attributes
10.  *Sample*[ ][ ]: array for original minority class samples
11.  *newindex*: keeps a count of number of synthetic samples generated, initialized to 0
12.  *Synthetic*[ ][ ]: array for synthetic samples
    (* *Compute k nearest neighbors for each minority class sample only.* *)
13.  **for** $i \leftarrow 1$ **to** $T$
14.    Compute $k$ nearest neighbors for $i$, and save the indices in the *nnarray*
15.    Populate($N$, $i$, *nnarray*)
16.  **endfor**

    *Populate*($N$, $i$, *nnarray*) (* *Function to generate the synthetic samples.* *)

17.  **while** $N \neq 0$
18.    Choose a random number between 1 and $k$, call it $nn$. This step chooses one of
    the $k$ nearest neighbors of $i$.
19.    **for** $attr \leftarrow 1$ **to** *numattrs*
20.        Compute: $dif = Sample[nnarray[nn]][attr] - Sample[i][attr]$
21.        Compute: $gap$ = random number between 0 and 1
22.        $Synthetic[newindex][attr] = Sample[i][attr] + gap * dif$
23.    **endfor**
24.    $newindex++$
25.    $N = N - 1$
26.  **endwhile**
27.  **return** (* *End of Populate.* *)
    End of Pseudo-Code.

Consider a sample (6,4) and let (4,3) be its nearest neighbor.
(6,4) is the sample for which k-nearest neighbors are being identified.
(4,3) is one of its k-nearest neighbors.
Let:
f1_1 = 6  f2_1 = 4  f2_1 - f1_1 = -2
f1_2 = 4  f2_2 = 3  f2_2 - f1_2 = -1
The new samples will be generated as
(f1',f2') = (6,4) + rand(0-1) * (-2,-1)
rand(0-1) generates a random number between 0 and 1.

Table 1: Example of generation of synthetic examples (SMOTE).

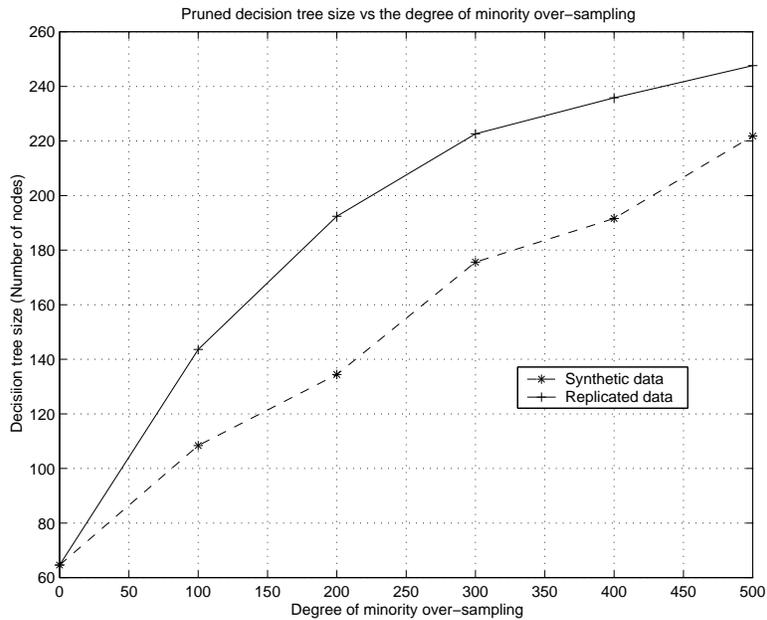

Figure 4: Comparison of decision tree sizes for replicated over-sampling and SMOTE for the Mammography dataset

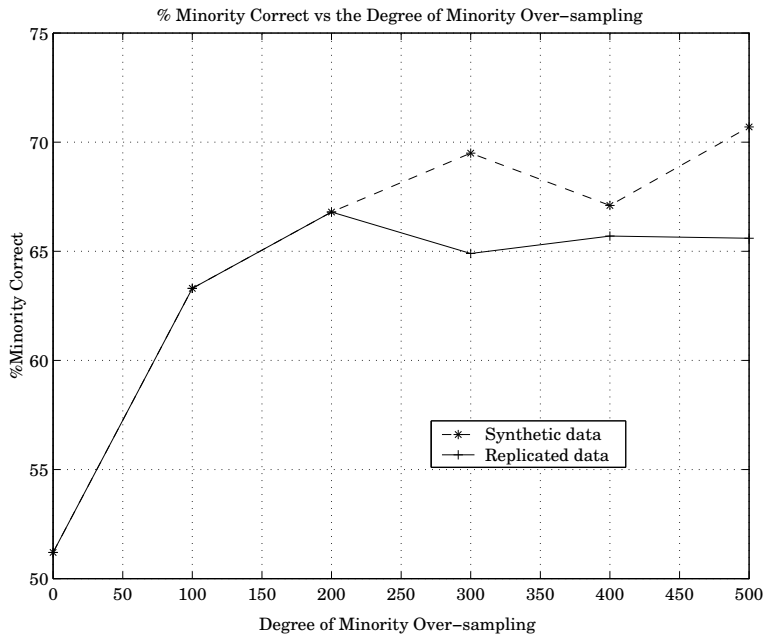

Figure 5: Comparison of % Minority correct for replicated over-sampling and SMOTE for the Mammography dataset

## 4.3 Under-sampling and SMOTE Combination

The majority class is under-sampled by randomly removing samples from the majority class population until the minority class becomes some specified percentage of the majority class. This forces the learner to experience varying degrees of under-sampling and at higher degrees of under-sampling the minority class has a larger presence in the training set. In describing our experiments, our terminology will be such that if we *under-sample the majority class at 200%*, it would mean that the modified dataset will contain *twice as many elements from the minority class as from the majority class*; that is, if the minority class had 50 samples and the majority class had 200 samples and we under-sample majority at 200%, the majority class would end up having 25 samples. By applying a combination of under-sampling and over-sampling, the initial bias of the learner towards the negative (majority) class is reversed in the favor of the positive (minority) class. Classifiers are learned on the dataset perturbed by "SMOTING" the minority class and under-sampling the majority class.

## 5. Experiments

We used three different machine learning algorithms for our experiments. Figure 6 provides an overview of our experiments.

1. **C4.5:** We compared various combinations of SMOTE and under-sampling with plain under-sampling using C4.5 release 8 (Quinlan, 1992) as the base classifier.

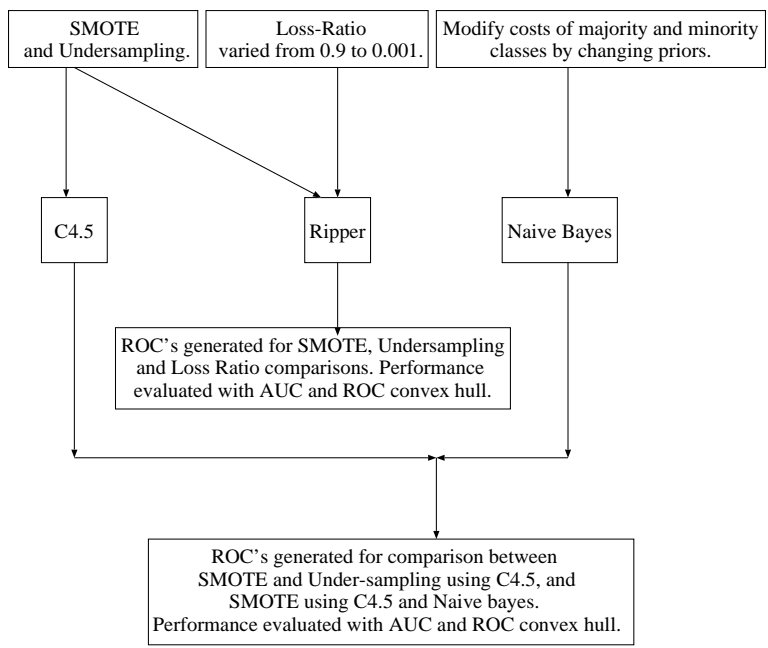

Figure 6: Experiments Overview

2. **Ripper:** We compared various combinations of SMOTE and under-sampling with plain under-sampling using Ripper (Cohen, 1995b) as the base classifier. We also varied Ripper's loss ratio (Cohen & Singer, 1996; Lewis & Catlett, 1994) from 0.9 to 0.001 (as a means of varying misclassification cost) and compared the effect of this variation with the combination of SMOTE and under-sampling. By reducing the loss ratio from 0.9 to 0.001 we were able to build a set of rules for the minority class.

3. **Naive Bayes Classifier:** The Naive Bayes Classifier[2] can be made cost-sensitive by varying the priors of the minority class. We varied the priors of the minority class from 1 to 50 times the majority class and compared with C4.5's SMOTE and under-sampling combination.

These different learning algorithms allowed SMOTE to be compared to some methods that can handle misclassification costs directly. %FP and %TP were averaged over 10-fold cross-validation runs for each of the data combinations. The minority class examples were over-sampled by calculating the five nearest neighbors and generating synthetic examples. The AUC was calculated using the trapezoidal rule. We extrapolated an extra point of TP = 100% and FP = 100% for each ROC curve. We also computed the ROC convex hull to identify the optimal classifiers, as the points lying on the hull are potentially optimal classifiers (Provost & Fawcett, 2001).

---

2. The source code was downloaded from http://fuzzy.cs.uni-magdeburg.de/~borgelt/software.html.

## 5.1 Datasets

We experimented on nine different datasets. These datasets are summarized in Table 5.2. These datasets vary extensively in their size and class proportions, thus offering different domains for SMOTE. In order of increasing imbalance they are:

1. The Pima Indian Diabetes (Blake & Merz, 1998) has 2 classes and 768 samples. The data is used to identify the positive diabetes cases in a population near Phoenix, Arizona. The number of positive class samples is only 268. Good sensitivity to detection of diabetes cases will be a desirable attribute of the classifier.

2. The Phoneme dataset is from the ELENA project[3]. The aim of the dataset is to distinguish between nasal (class 0) and oral sounds (class 1). There are 5 features. The class distribution is 3,818 samples in class 0 and 1,586 samples in class 1.

3. The Adult dataset (Blake & Merz, 1998) has 48,842 samples with 11,687 samples belonging to the minority class. This dataset has 6 continuous features and 8 nominal features. SMOTE and SMOTE-NC (see Section 6.1) algorithms were evaluated on this dataset. For SMOTE, we extracted the continuous features and generated a new dataset with only continuous features.

4. The E-state data[4] (Hall, Mohney, & Kier, 1991) consists of electrotopological state descriptors for a series of compounds from the National Cancer Institute's Yeast Anti-Cancer drug screen. E-state descriptors from the NCI Yeast AntiCancer Drug Screen were generated by Tripos, Inc. Briefly, a series of about 60,000 compounds were tested against a series of 6 yeast strains at a given concentration. The test was a high-throughput screen at only one concentration so the results are subject to contamination, etc. The growth inhibition of the yeast strain when exposed to the given compound (with respect to growth of the yeast in a neutral solvent) was measured. The activity classes are either active — at least one single yeast strain was inhibited more than 70%, or inactive — no yeast strain was inhibited more than 70%. The dataset has 53,220 samples with 6,351 samples of active compounds.

5. The Satimage dataset (Blake & Merz, 1998) has 6 classes originally. We chose the smallest class as the minority class and collapsed the rest of the classes into one as was done in (Provost et al., 1998). This gave us a skewed 2-class dataset, with 5809 majority class samples and 626 minority class samples.

6. The Forest Cover dataset is from the UCI repository (Blake & Merz, 1998). This dataset has 7 classes and 581,012 samples. This dataset is for the prediction of forest cover type based on cartographic variables. Since our system currently works for binary classes we extracted data for two classes from this dataset and ignored the rest. Most other approaches only work for only two classes (Ling & Li, 1998; Japkowicz, 2000; Kubat & Matwin, 1997; Provost & Fawcett, 2001). The two classes we considered are Ponderosa Pine with 35,754 samples and Cottonwood/Willow with 2,747 samples. Nevertheless, the SMOTE technique can be applied to a multiple class problem as well by specifying what class to SMOTE for. However, in this paper, we have focused on 2-classes problems, to explicitly represent positive and negative classes.

---

3. ftp.dice.ucl.ac.be in the directory pub/neural-nets/ELENA/databases.
4. We would like to thank Steven Eschrich for providing the dataset and description to us.

| Dataset | Majority Class | Minority Class |
|---|---|---|
| Pima | 500 | 268 |
| Phoneme | 3818 | 1586 |
| Adult | 37155 | 11687 |
| E-state | 46869 | 6351 |
| Satimage | 5809 | 626 |
| Forest Cover | 35754 | 2747 |
| Oil | 896 | 41 |
| Mammography | 10923 | 260 |
| Can | 435512 | 8360 |

Table 2: Dataset distribution

7. The Oil dataset was provided by Robert Holte and is used in their paper (Kubat et al., 1998). This dataset has 41 oil slick samples and 896 non-oil slick samples.

8. The Mammography dataset (Woods et al., 1993) has 11,183 samples with 260 calcifications. If we look at predictive accuracy as a measure of goodness of the classifier for this case, the default accuracy would be 97.68% when every sample is labeled non-calcification. But, it is desirable for the classifier to predict most of the calcifications correctly.

9. The Can dataset was generated from the Can ExodusII data using the AVATAR (Chawla & Hall, 1999) version of the Mustafa Visualization tool[5]. The portion of the can being crushed was marked as "very interesting" and the rest of the can was marked as "unknown." A dataset of size 443,872 samples with 8,360 samples marked as "very interesting" was generated.

## 5.2 ROC Creation

A ROC curve for SMOTE is produced by using C4.5 or Ripper to create a classifier for each one of a series of modified training datasets. A given ROC curve is produced by first over-sampling the minority class to a specified degree and then under-sampling the majority class at increasing degrees to generate the successive points on the curve. The amount of under-sampling is identical to plain under-sampling. So, each corresponding point on each ROC curve for a dataset represents the same number of majority class samples. Different ROC curves are produced by starting with different levels of minority over-sampling. ROC curves were also generated by varying the loss ratio in Ripper from 0.9 to 0.001 and by varying the priors of the minority class from the original distribution to up to 50 times the majority class for a Naive Bayes Classifier.

---

5. The Mustafa visualization tool was developed by Mike Glass of Sandia National Labs.

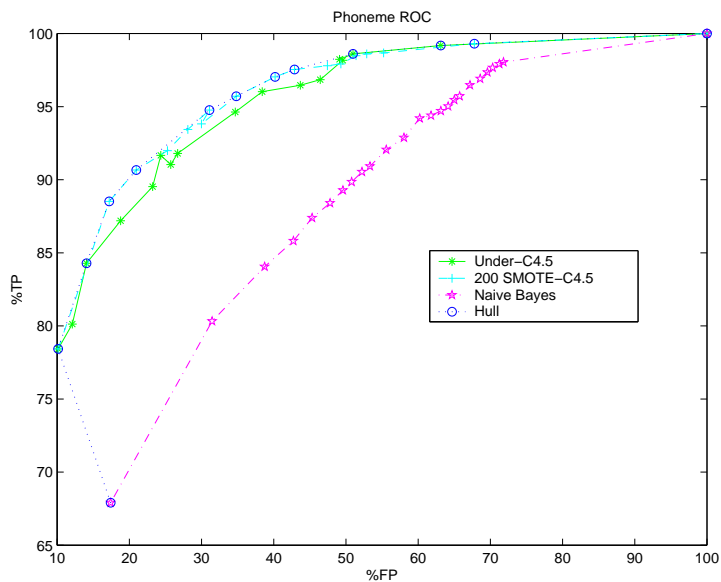

Figure 7: Phoneme. Comparison of SMOTE-C4.5, Under-C4.5, and Naive Bayes. SMOTE-C4.5 dominates over Naive Bayes and Under-C4.5 in the ROC space. SMOTE-C4.5 classifiers are potentially optimal classifiers.

Figures 9 through 23 show the experimental ROC curves obtained for the nine datasets with the three classifiers. The ROC curve for plain under-sampling of the majority class (Ling & Li, 1998; Japkowicz, 2000; Kubat & Matwin, 1997; Provost & Fawcett, 2001) is compared with our approach of combining synthetic minority class over-sampling (SMOTE) with majority class under-sampling. The plain under-sampling curve is labeled "Under", and the SMOTE and under-sampling combination ROC curve is labeled "SMOTE". Depending on the size and relative imbalance of the dataset, one to five SMOTE and under-sampling curves are created. We only show the best results from SMOTE combined with under-sampling and the plain under-sampling curve in the graphs. The SMOTE ROC curve from C4.5 is also compared with the ROC curve obtained from varying the priors of minority class using a Naive Bayes classifier — labeled as "Naive Bayes". "SMOTE", "Under", and "Loss Ratio" ROC curves, generated using Ripper are also compared. For a given family of ROC curves, an ROC convex hull (Provost & Fawcett, 2001) is generated. The ROC convex hull is generated using the Graham's algorithm (O'Rourke, 1998). For reference, we show the ROC curve that would be obtained using minority over-sampling by replication in Figure 19.

Each point on the ROC curve is the result of either a classifier (C4.5 or Ripper) learned for a particular combination of under-sampling and SMOTE, a classifier (C4.5 or Ripper) learned with plain under-sampling, or a classifier (Ripper) learned using some loss ratio or a classifier (Naive Bayes) learned for a different prior for the minority class. Each point represents the average (%TP and %FP) 10-fold cross-validation result. The lower leftmost point for a given ROC curve is from the raw dataset, without any majority class under-sampling or minority class over-sampling. The minority class was over-sampled at 50%, 100%, 200%, 300%, 400%, 500%. The majority class was under-sampled at 10%, 15%, 25%, 50%, 75%, 100%, 125%, 150%, 175%, 200%, 300%, 400%, 500%, 600%, 700%, 800%, 1000%, and 2000%. The amount of majority class under-sampling and minority class over-sampling depended on the dataset size and class proportions. For instance, consider the ROC curves in Figure 17 for the mammography dataset. There are three curves — one for plain majority class under-sampling in which the range of under-sampling is varied between 5% and 2000% at different intervals, one for a combination of SMOTE and majority class under-sampling, and one for Naive Bayes — and one ROC convex hull curve. The ROC curve shown in Figure 17 is for the minority class over-sampled at 400%. Each point on the SMOTE ROC curves represents a combination of (synthetic) over-sampling and under-sampling, the amount of under-sampling follows the same range as for plain under-sampling. For a better understanding of the ROC graphs, we have shown different sets of ROC curves for one of our datasets in Appendix A.

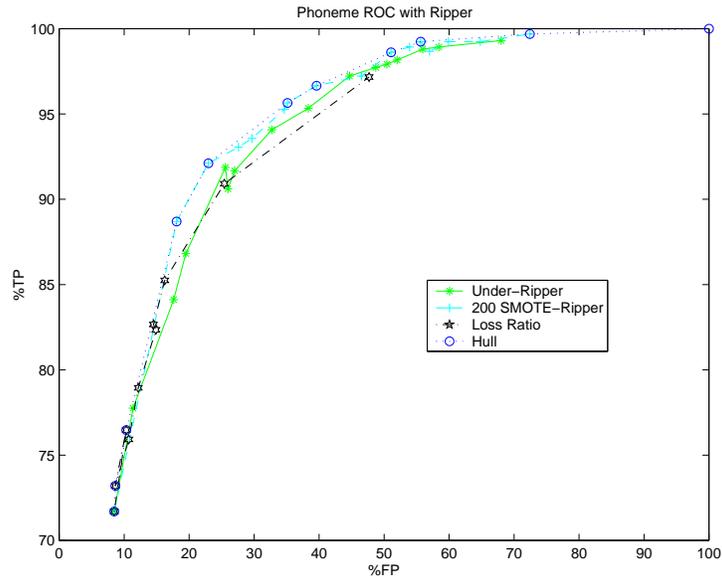

Figure 8: Phoneme. Comparison of SMOTE-Ripper, Under-Ripper, and modifying Loss Ratio in Ripper. SMOTE-Ripper dominates over Under-Ripper and Loss Ratio in the ROC space. More SMOTE-Ripper classifiers lie on the ROC convex hull.

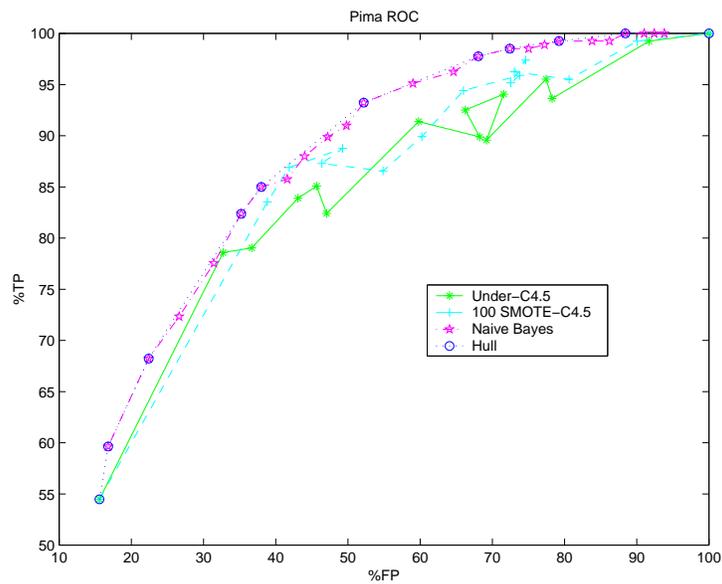

Figure 9: Pima Indians Diabetes. Comparison of SMOTE-C4.5, Under-C4.5, and Naive Bayes. Naive Bayes dominates over SMOTE-C4.5 in the ROC space.

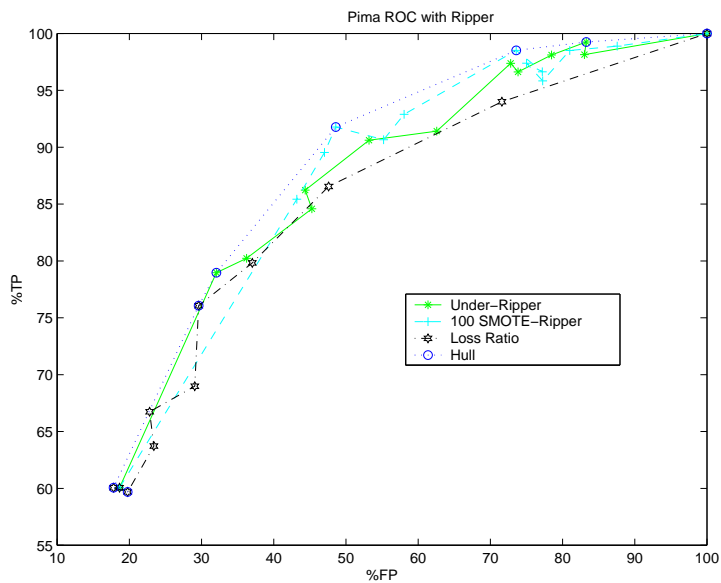

Figure 10: Pima Indians Diabetes. Comparison of SMOTE-Ripper, Under-Ripper, and modifying Loss Ratio in Ripper. SMOTE-Ripper dominates over Under-Ripper and Loss Ratio in the ROC space.

For the Can dataset, we had to SMOTE to a lesser degree than for the other datasets due to the structural nature of the dataset. For the Can dataset there is a structural neighborhood already established in the mesh geometry, so SMOTE can lead to creating neighbors which are under the surface (and hence not interesting), since we are looking at the feature space of physics variables and not the structural information.

The ROC curves show a trend that as we increase the amount of under-sampling coupled with over-sampling, our minority classification accuracy increases, of course at the expense of more majority class errors. For almost all the ROC curves, the SMOTE approach dom-inates. Adhering to the definition of ROC convex hull, most of the potentially optimal classifiers are the ones generated with SMOTE.

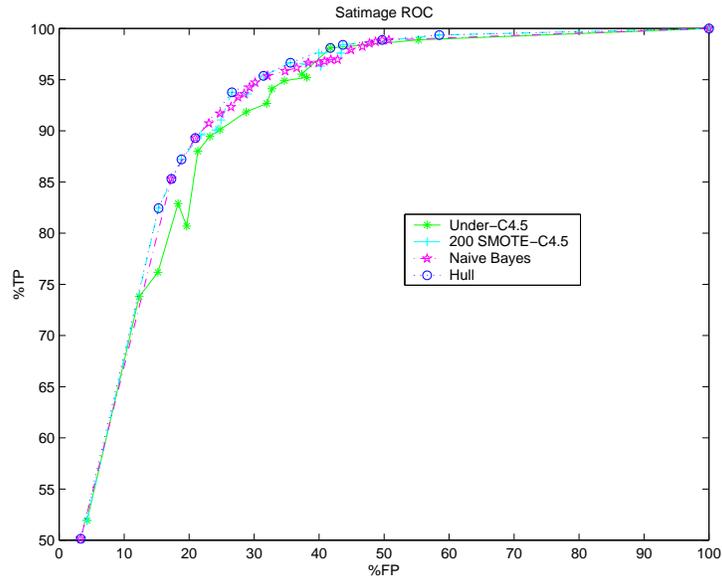

Figure 11: Satimage. Comparison of SMOTE-C4.5, Under-C4.5, and Naive Bayes. The ROC curves of Naive Bayes and SMOTE-C4.5 show an overlap; however, at higher TP's more points from SMOTE-C4.5 lie on the ROC convex hull.

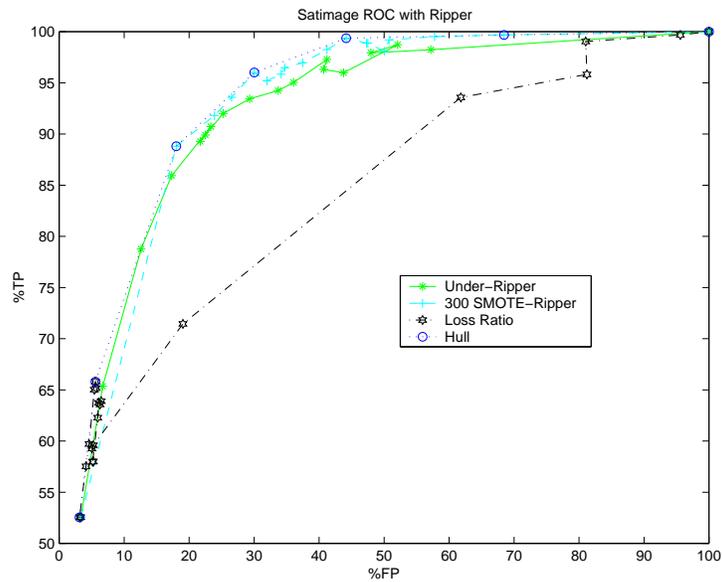

Figure 12: Satimage. Comparison of SMOTE-Ripper, Under-Ripper, and modifying Loss Ratio in Ripper. SMOTE-Ripper dominates the ROC space. The ROC convex hull is mostly constructed with points from SMOTE-Ripper.

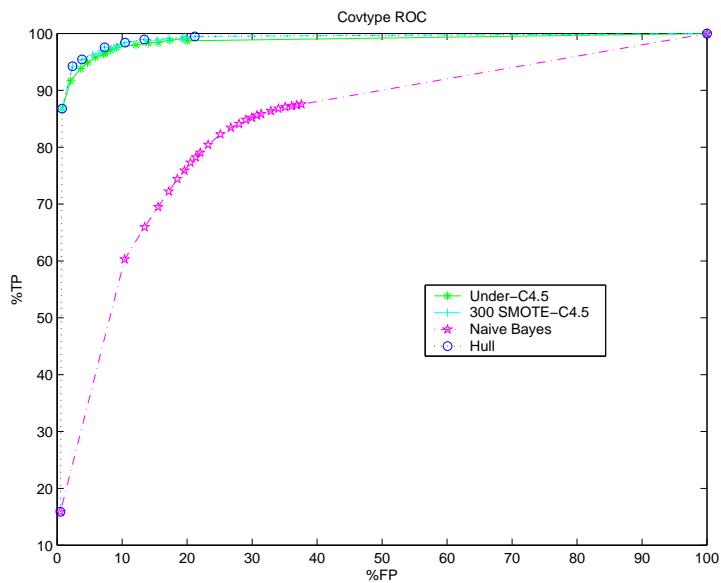

Figure 13: Forest Cover. Comparison of SMOTE-C4.5, Under-C4.5, and Naive Bayes. SMOTE-C4.5 and Under-C4.5 ROC curves are very close to each other. However, more points from the SMOTE-C4.5 ROC curve lie on the ROC convex hull, thus establishing a dominance.

## 5.3 AUC Calculation

The Area Under the ROC curve (AUC) is calculated using a form of the trapezoid rule. The lower leftmost point for a given ROC curve is a classifier's performance on the raw data. The upper rightmost point is always (100%, 100%). If the curve does not naturally end at this point, the point is added. This is necessary in order for the AUC's to be compared over the same range of %FP.

The AUCs listed in Table 5.3 show that for all datasets the combined synthetic minority over-sampling and majority over-sampling is able to improve over plain majority under-sampling with C4.5 as the base classifier. Thus, our SMOTE approach provides an improvement in correct classification of data in the underrepresented class. The same conclusion holds from an examination of the ROC convex hulls. Some of the entries are missing in the table, as SMOTE was not applied at the same amounts to all datasets. The amount of SMOTE was less for less skewed datasets. Also, we have not included AUC's for Ripper/Naive Bayes. The ROC convex hull identifies SMOTE classifiers to be potentially optimal as compared to plain under-sampling or other treatments of misclassification costs, generally. Exceptions are as follows: for the Pima dataset, Naive Bayes dominates over SMOTE-C4.5; for the Oil dataset, Under-Ripper dominates over SMOTE-Ripper. For the Can dataset, SMOTE-*classifier* (*classifier* = C4.5 or Ripper) and Under-*classifier* ROC

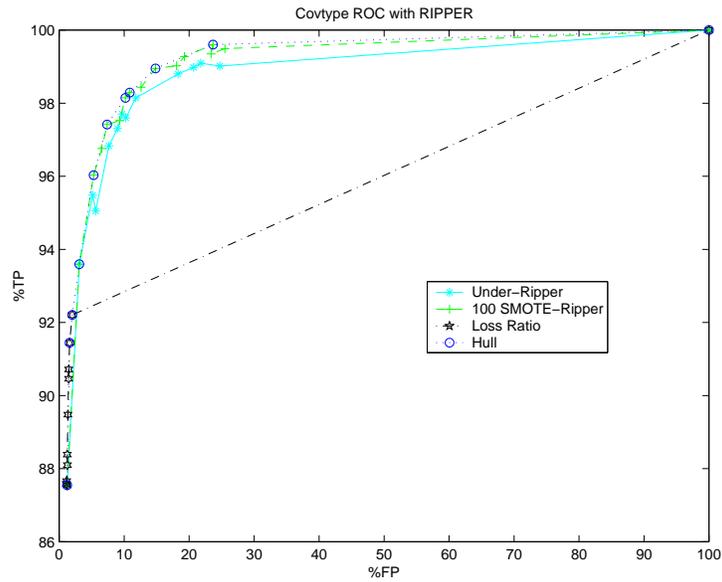

Figure 14: Forest Cover. Comparison of SMOTE-Ripper, Under-Ripper, and modifying Loss Ratio in Ripper. SMOTE-Ripper shows a domination in the ROC space. More points from SMOTE-Ripper curve lie on the ROC convex hull.

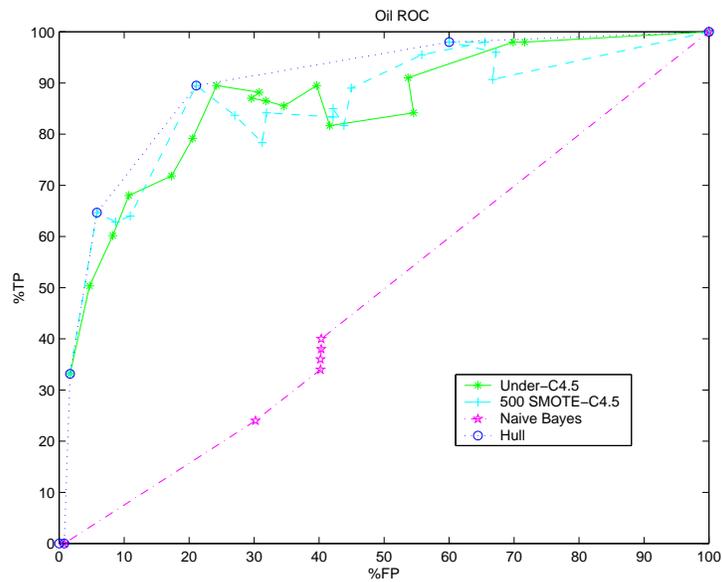

Figure 15: Oil. Comparison of SMOTE-C4.5, Under-C4.5, and Naive Bayes. Although, SMOTE-C4.5 and Under-C4.5 ROC curves intersect at points, more points from SMOTE-C4.5 curve lie on the ROC convex hull.

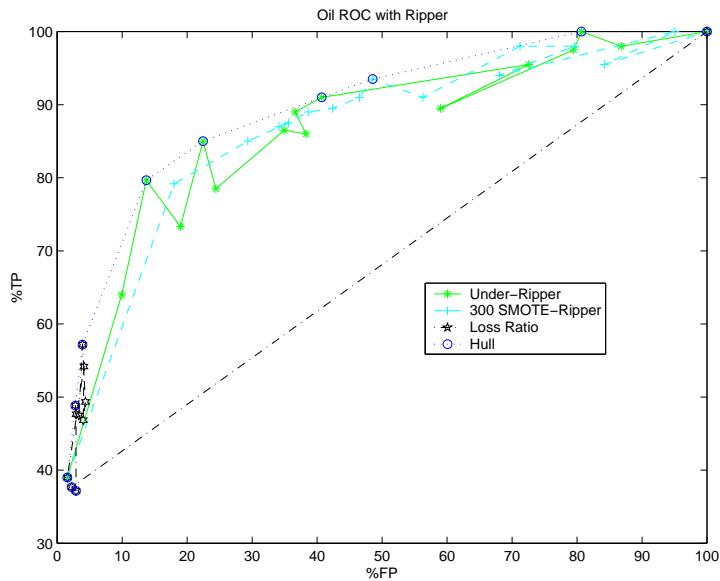

Figure 16: Oil. Comparison of SMOTE-Ripper, Under-Ripper, and modifying Loss Ratio in Ripper. Under-Ripper and SMOTE-Ripper curves intersect, and more points from the Under-Ripper curve lie on the ROC convex hull.

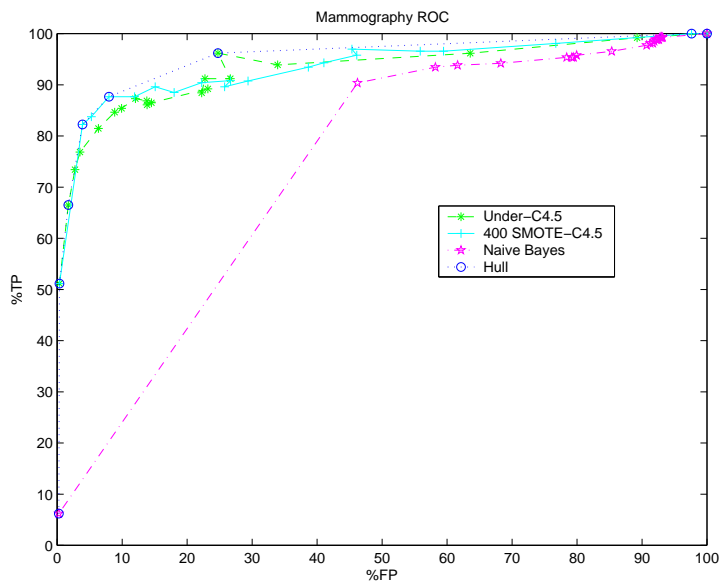

Figure 17: Mammography. Comparison of SMOTE-C4.5, Under-C4.5, and Naive Bayes. SMOTE-C4.5 and Under-C4.5 curves intersect in the ROC space; however, by virtue of number of points on the ROC convex hull, SMOTE-C4.5 has more potentially optimal classifiers.

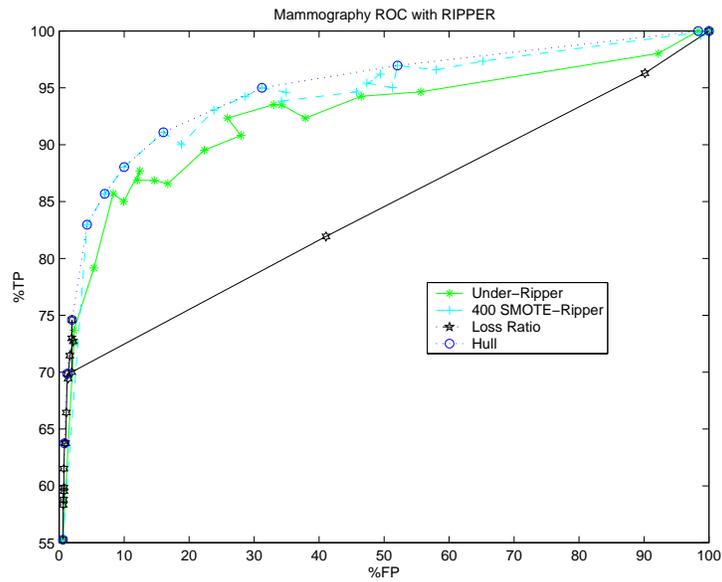

Figure 18: Mammography. Comparison of SMOTE-Ripper, Under-Ripper, and modifying Loss Ratio in Ripper. SMOTE-Ripper dominates the ROC space for TP > 75%.

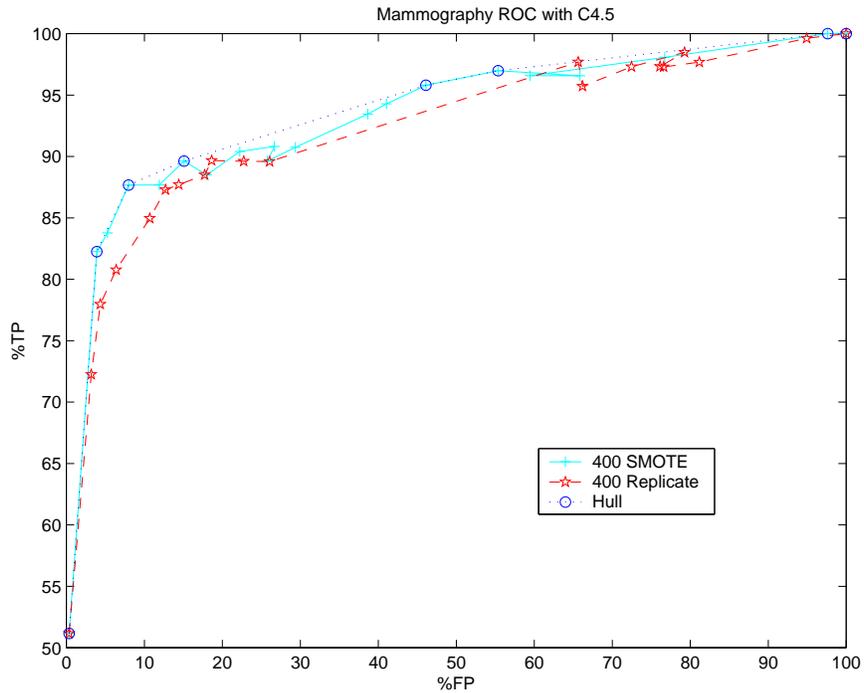

Figure 19: A comparison of over-sampling minority class examples by SMOTE and over-sampling the minority class examples by replication for the Mammography dataset.

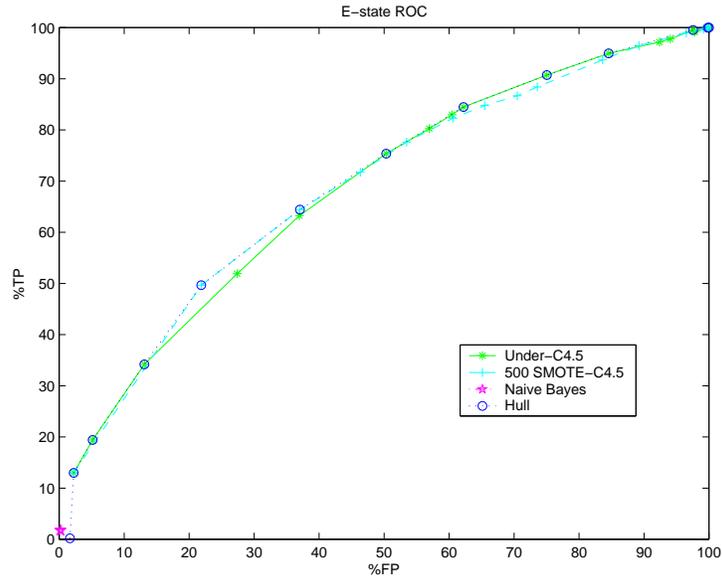

Figure 20: E-state. (a) Comparison of SMOTE-C4.5, Under-C4.5, and Naive Bayes. SMOTE-C4.5 and Under-C4.5 curves intersect in the ROC space; however, SMOTE-C4.5 has more potentially optimal classifiers, based on the number of points on the ROC convex hull.

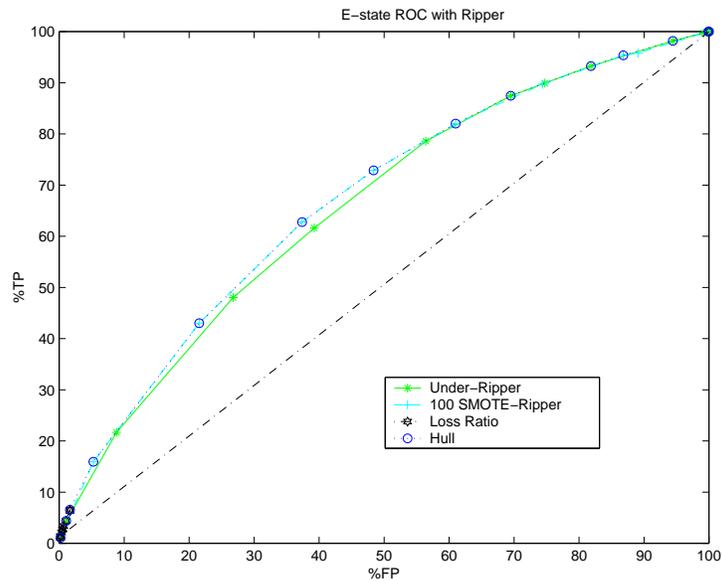

Figure 21: E-state. Comparison of SMOTE-Ripper, Under-Ripper, and modifying Loss Ratio in Ripper. SMOTE-Ripper has more potentially optimal classifiers, based on the number of points on the ROC convex hull.

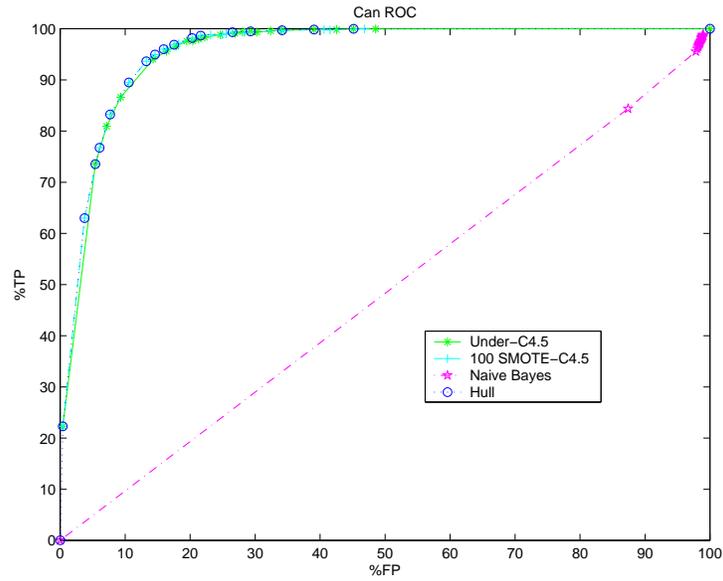

Figure 22: Can. Comparison of SMOTE-C4.5, Under-C4.5, and Naive Bayes. SMOTE-C4.5 and Under-C4.5 ROC curves overlap for most of the ROC space.

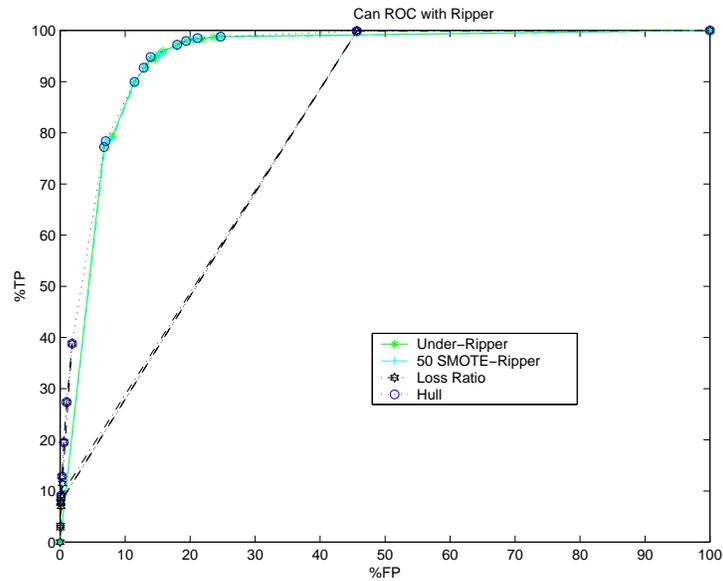

Figure 23: Can. Comparison of SMOTE-Ripper, Under-Ripper, and modifying Loss Ratio in Ripper. SMOTE-Ripper and Under-Ripper ROC curves overlap for most of the ROC space.

| Dataset | Under | 50 SMOTE | 100 SMOTE | 200 SMOTE | 300 SMOTE | 400 SMOTE | 500 SMOTE |
|---|---|---|---|---|---|---|---|
| Pima | 7242 | | **7307** | | | | |
| Phoneme | 8622 | | 8644 | **8661** | | | |
| Satimage | 8900 | | 8957 | **8979** | 8963 | 8975 | 8960 |
| Forest Cover | 9807 | | 9832 | 9834 | **9849** | 9841 | 9842 |
| Oil | 8524 | | 8523 | 8368 | 8161 | 8339 | **8537** |
| Mammography | 9260 | | 9250 | 9265 | 9311 | **9330** | 9304 |
| E-state | 6811 | | 6792 | **6828** | 6784 | 6788 | 6779 |
| Can | 9535 | **9560** | 9505 | 9505 | 9494 | 9472 | 9470 |

Table 3: AUC's [C4.5 as the base classifier] with the best highlighted in bold.

curves overlap in the ROC space. For all the other datasets, SMOTE-*classifier* has more potentially optimal classifiers than any other approach.

## 5.4 Additional comparison to changing the decision thresholds

Provost (2000) suggested that simply changing the decision threshold should always be considered as an alternative to more sophisticated approaches. In the case of C4.5, this would mean changing the decision threshold at the leaves of the decision trees. For example, a leaf could classify examples as the minority class even if more than 50% of the training examples at the leaf represent the majority class. We experimented by setting the decision thresholds at the leaves for the C4.5 decision tree learner at 0.5, 0.45, 0.42, 0.4, 0.35, 0.32, 0.3, 0.27, 0.25, 0.22, 0.2, 0.17, 0.15, 0.12, 0.1, 0.05, 0.0. We experimented on the Phoneme dataset. Figure 24 shows the comparison of the SMOTE and under-sampling combination against C4.5 learning by tuning the bias towards the minority class. The graph shows that the SMOTE and under-sampling combination ROC curve is dominating over the entire range of values.

## 5.5 Additional comparison to one-sided selection and SHRINK

For the oil dataset, we also followed a slightly different line of experiments to obtain results comparable to (Kubat et al., 1998). To alleviate the problem of imbalanced datasets the authors have proposed (a) one-sided selection for under-sampling the majority class (Kubat & Matwin, 1997) and (b) the SHRINK system (Kubat et al., 1998). Table 5.5 contains the results from (Kubat et al., 1998). Acc+ is the accuracy on positive (minority) examples and Acc− is the accuracy on the negative (majority) examples. Figure 25 shows the trend for Acc+ and Acc− for one combination of the SMOTE strategy and varying degrees of under-sampling of the majority class. The Y-axis represents the accuracy and the X-axis represents the percentage majority class under-sampled. The graphs indicate that in the band of under-sampling between 50% and 125% the results are comparable to those achieved by SHRINK and better than SHRINK in some cases. Table 5.5 summarizes the results for the SMOTE at 500% and under-sampling combination. We also tried combinations of SMOTE at 100-400% and varying degrees of under-sampling and achieved comparable results. The

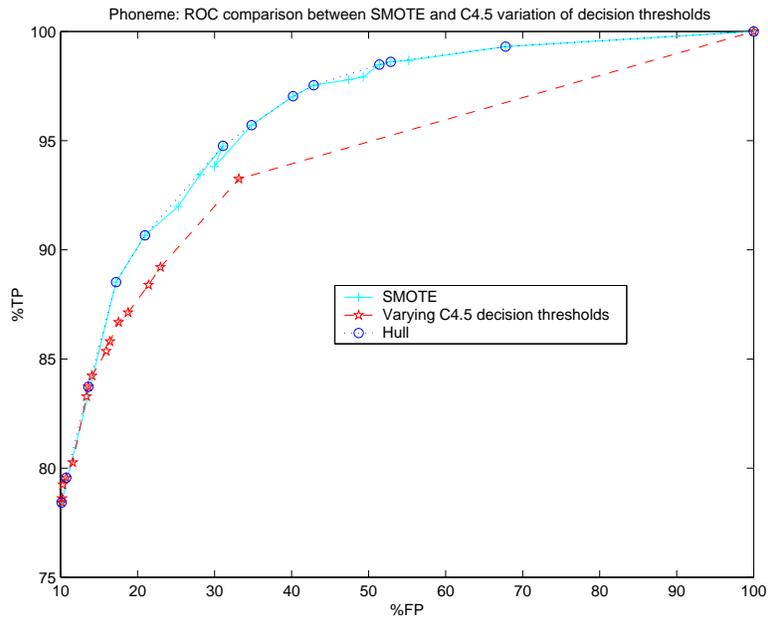

Figure 24: SMOTE and Under-sampling combination against C4.5 learning by tuning the bias towards the minority class

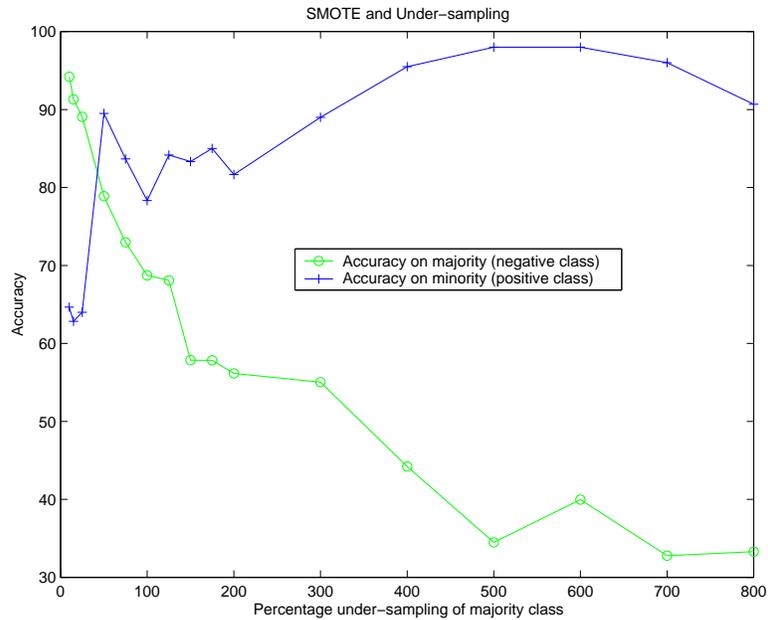

Figure 25: SMOTE (500 OU) and Under-sampling combination performance

SHRINK approach and our SMOTE approach are not directly comparable, though, as they see different data points. SMOTE offers no clear improvement over one-sided selection.

| Method | Acc+ | Acc− |
|---|---|---|
| SHRINK | 82.5% | 60.9% |
| One-sided selection | 76.0% | 86.6% |

Table 4: Cross-validation results (Kubat et al., 1998)

| Under-sampling % | Acc+ | Acc− |
|---|---|---|
| 10% | 64.7% | 94.2% |
| 15% | 62.8% | 91.3% |
| 25% | 64.0% | 89.1% |
| 50% | 89.5% | 78.9% |
| 75% | 83.7% | 73.0% |
| 100% | 78.3% | 68.7% |
| 125% | 84.2% | 68.1% |
| 150% | 83.3% | 57.8% |
| 175% | 85.0% | 57.8% |
| 200% | 81.7% | 56.7% |
| 300% | 89.0% | 55.0% |
| 400% | 95.5% | 44.2% |
| 500% | 98.0% | 35.5% |
| 600% | 98.0% | 40.0% |
| 700% | 96.0% | 32.8% |
| 800% | 90.7% | 33.3% |

Table 5: Cross-validation results for SMOTE at 500% SMOTE on the Oil data set.

## 6. Future Work

There are several topics to be considered further in this line of research. Automated adaptive selection of the number of nearest neighbors would be valuable. Different strategies for creating the synthetic neighbors may be able to improve the performance. Also, selecting nearest neighbors with a focus on examples that are incorrectly classified may improve performance. A minority class sample could possibly have a majority class sample as its nearest neighbor rather than a minority class sample. This crowding will likely contribute to the redrawing of the decision surfaces in favor of the minority class. In addition to these topics, the following subsections discuss two possible extensions of SMOTE, and an application of SMOTE to information retrieval.

### 6.1 SMOTE-NC

While our SMOTE approach currently does not handle data sets with all nominal features, it was generalized to handle mixed datasets of continuous and nominal features. We call this approach Synthetic Minority Over-sampling TEchnique-Nominal Continuous [SMOTE-NC]. We tested this approach on the Adult dataset from the UCI repository. The SMOTE-NC algorithm is described below.

1. Median computation: Compute the median of standard deviations of all continuous features for the minority class. If the nominal features differ between a sample and its potential nearest neighbors, then this median is included in the Euclidean distance computation. We use median to penalize the difference of nominal features by an amount that is related to the typical difference in continuous feature values.

2. Nearest neighbor computation: Compute the Euclidean distance between the feature vector for which k-nearest neighbors are being identified (minority class sample) and the other feature vectors (minority class samples) using the continuous feature space. For every differing nominal feature between the considered feature vector and its potential nearest-neighbor, include the median of the standard deviations previously computed, in the Euclidean distance computation. Table 2 demonstrates an example.

---

F1 = 1 2 3 A B C [Let this be the sample for which we are computing nearest neighbors]

F2 = 4 6 5 A D E

F3 = 3 5 6 A B K

So, Euclidean Distance between F2 and F1 would be:

Eucl = sqrt[$(4-1)^2 + (6-2)^2 + (5-3)^2 + \text{Med}^2 + \text{Med}^2$]

**Med** is the median of the standard deviations of continuous features of the minority class.

The median term is included twice for feature numbers 5: B→D and 6: C→E, which differ for the two feature vectors: F1 and F2.

---

Table 6: Example of nearest neighbor computation for SMOTE-NC.

3. Populate the synthetic sample: The continuous features of the new synthetic minority class sample are created using the same approach of SMOTE as described earlier. The nominal feature is given the value occuring in the majority of the k-nearest neighbors.

The SMOTE-NC experiments reported here are set up the same as those with SMOTE, except for the fact that we examine one dataset only. SMOTE-NC with the Adult dataset differs from our typical result: it performs worse than plain under-sampling based on AUC, as shown in Figures 26 and 27. We extracted only continuous features to separate the effect of SMOTE and SMOTE-NC on this dataset, and to determine whether this oddity was due to our handling of nominal features. As shown in Figure 28, even SMOTE with only continuous features applied to the Adult dataset, does not achieve any better performance than plain under-sampling. Some of the minority class continuous features have a very high variance, so, the synthetic generation of minority class samples could be overlapping with the majority class space, thus leading to more false positives than plain under-sampling. This hypothesis is also supported by the decreased AUC measure as we SMOTE at degrees greater than 50%. The higher degrees of SMOTE lead to more minority class samples in the dataset, and thus a greater overlap with the majority class decision space.

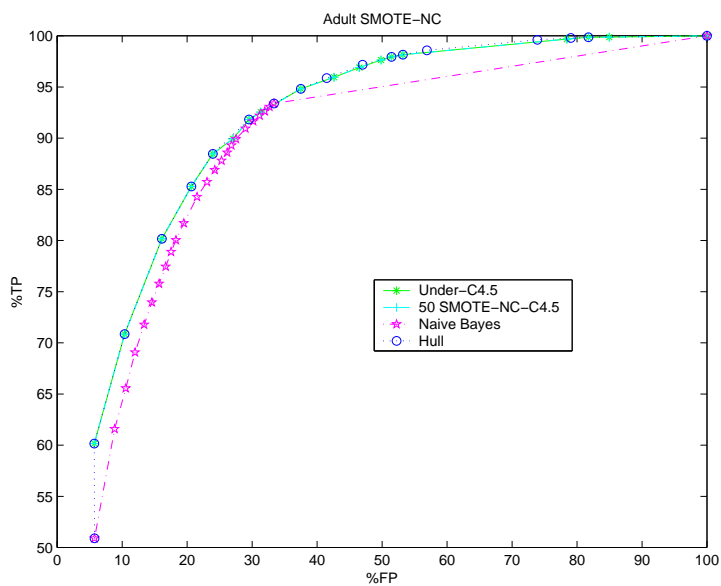

Figure 26: Adult. Comparison of SMOTE-C4.5, Under-C4.5, and Naive Bayes. SMOTE-C4.5 and Under-C4.5 ROC curves overlap for most of the ROC space.

## 6.2 SMOTE-N

Potentially, SMOTE can also be extended for nominal features — SMOTE-N — with the nearest neighbors computed using the modified version of Value Difference Metric (Stanfill & Waltz, 1986) proposed by Cost and Salzberg (1993). The Value Difference Metric (VDM) looks at the overlap of feature values over all feature vectors. A matrix defining the distance between corresponding feature values for all feature vectors is created. The distance $\delta$ between two corresponding feature values is defined as follows.

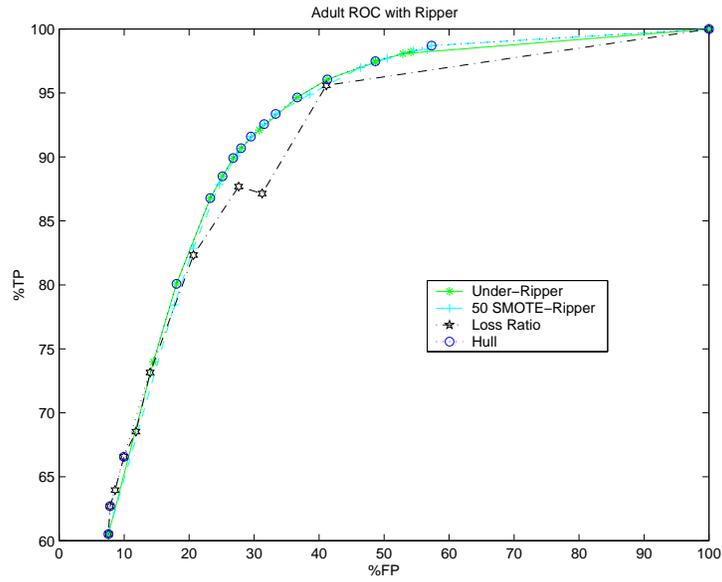

Figure 27: Adult. Comparison of SMOTE-Ripper, Under-Ripper, and modifying Loss Ratio in Ripper. SMOTE-Ripper and Under-Ripper ROC curves overlap for most of the ROC space.

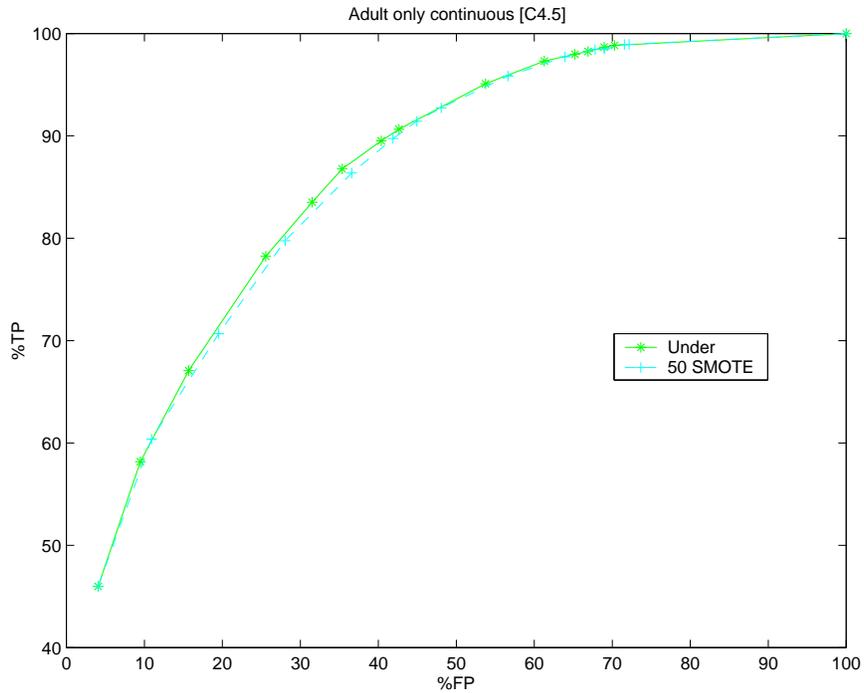

Figure 28: Adult with only continuous features. The overlap of SMOTE-C4.5 and Under-C4.5 is observed under this scenario as well.

$$\delta(V_1, V_2) = \sum_{i=1}^{n} |\frac{C_{1i}}{C_1} - \frac{C_{2i}}{C_2}|^k \tag{1}$$

In the above equation, $V_1$ and $V_2$ are the two corresponding feature values. $C_1$ is the total number of occurrences of feature value $V_1$, and $C_{1i}$ is the number of occurrences of feature value $V_1$ for class $i$. A similar convention can also be applied to $C_{2i}$ and $C_2$. $k$ is a constant, usually set to 1. This equation is used to compute the matrix of value differences for each nominal feature in the given set of feature vectors. Equation 1 gives a geometric distance on a fixed, finite set of values (Cost & Salzberg, 1993). Cost and Salzberg's modified VDM omits the weight term $w_f^a$ included in the $\delta$ computation by Stanfill and Waltz, which has an effect of making $\delta$ symmetric. The distance $\Delta$ between two feature vectors is given by:

$$\Delta(X, Y) = w_x w_y \sum_{i=1}^{N} \delta(x_i, y_i)^r \tag{2}$$

$r = 1$ yields the Manhattan distance, and $r = 2$ yields the Euclidean distance (Cost & Salzberg, 1993). $w_x$ and $w_y$ are the exemplar weights in the modified VDM. $w_y = 1$ for a new example (feature vector), and $w_x$ is the bias towards more reliable examples (feature vectors) and is computed as the ratio of the number of uses of a feature vector to the number of correct uses of the feature vector; thus, more accurate feature vectors will have $w_x \approx 1$. For SMOTE-N we can ignore these weights in equation 2, as SMOTE-N is not used for classification purposes directly. However, we can redefine these weights to give more weight to the minority class feature vectors falling closer to the majority class feature vectors; thus, making those minority class features appear further away from the feature vector under consideration. Since, we are more interested in forming broader but accurate regions of the minority class, the weights might be used to avoid populating along neighbors which fall closer to the majority class. To generate new minority class feature vectors, we can create new set feature values by taking the majority vote of the feature vector in consideration and its $k$ nearest neighbors. Table 6.2 shows an example of creating a synthetic feature vector.

---

Let F1 = A B C D E be the feature vector under consideration and let its 2 nearest neighbors be

F2 = A F C G N

F3 = H B C D N

The application of SMOTE-N would create the following feature vector:

FS = A B C D N

---

Table 7: Example of SMOTE-N

## 6.3 Application of SMOTE to Information Retrieval

We are investigating the application of SMOTE to information retrieval (IR). The IR problems come with a plethora of features and potentially many categories. SMOTE would have to be applied in conjunction with a feature selection algorithm, after transforming the given document or web page in a bag-of-words format.

An interesting comparison to SMOTE would be the combination of Naive Bayes and *Odds ratio*. *Odds ratio* focuses on a target class, and ranks documents according to their relevance to the target or positive class. SMOTE also focuses on a target class by creating more examples of that class.

## 7. Summary

The results show that the SMOTE approach can improve the accuracy of classifiers for a minority class. SMOTE provides a new approach to over-sampling. The combination of SMOTE and under-sampling performs better than plain under-sampling. SMOTE was tested on a variety of datasets, with varying degrees of imbalance and varying amounts of data in the training set, thus providing a diverse testbed. The combination of SMOTE and under-sampling also performs better, based on domination in the ROC space, than varying loss ratios in Ripper or by varying the class priors in Naive Bayes Classifier: the methods that could directly handle the skewed class distribution. SMOTE forces focused learning and introduces a bias towards the minority class. Only for Pima — the least skewed dataset — does the Naive Bayes Classifier perform better than SMOTE-C4.5. Also, only for the Oil dataset does the Under-Ripper perform better than SMOTE-Ripper. For the Can dataset, SMOTE-*classifier* and Under-*classifier* ROC curves overlap in the ROC space. For all the rest of the datasets SMOTE-*classifier* performs better than Under-*classifier*, Loss Ratio, and Naive Bayes. Out of a total of 48 experiments performed, SMOTE-*classifier* does not perform the best only for 4 experiments.

The interpretation of why synthetic minority over-sampling improves performance where as minority over-sampling with replacement does not is fairly straightforward. Consider the effect on the decision regions in feature space when minority over-sampling is done by replication (sampling with replacement) versus the introduction of synthetic examples. With replication, the decision region that results in a classification decision for the minority class can actually become smaller and more specific as the minority samples in the region are replicated. This is the opposite of the desired effect. Our method of synthetic over-sampling works to cause the classifier to build larger decision regions that contain nearby minority class points. The same reasons may be applicable to why SMOTE performs better than Ripper's loss ratio and Naive Bayes; these methods, nonetheless, are still learning from the information provided in the dataset, albeit with different cost information. SMOTE provides more related minority class samples to learn from, thus allowing a learner to carve broader decision regions, leading to more coverage of the minority class.

## Acknowledgments

This research was partially supported by the United States Department of Energy through the Sandia National Laboratories ASCI VIEWS Data Discovery Program, contract number

DE-AC04-76DO00789. We thank Robert Holte for providing the oil spill dataset used in their paper. We also thank Foster Provost for clarifying his method of using the Satimage dataset. We would also like to thank the anonymous reviewers for their various insightful comments and suggestions.

## Appendix A. ROC graphs for Oil Dataset

The following figures show different sets of ROC curves for the oil dataset. Figure 29 (a) shows the ROC curves for the Oil dataset, as included in the main text; Figure 29(b) shows the ROC curves without the ROC convex hull; Figure 29(c) shows the two convex hulls, obtained with and without SMOTE. The ROC convex hull shown by dashed lines and stars in Figure 29(c), was computed by including Under-C4.5 and Naive Bayes in the family of ROC curves. The ROC convex hull shown by solid line and small circles in Figure 29(c) was computed by including 500 SMOTE-C4.5, Under-C4.5, and Naive Bayes in the family of ROC curves. The ROC convex hull with SMOTE dominates the ROC convex hull without SMOTE, hence SMOTE-C4.5 contributes more optimal classifiers.

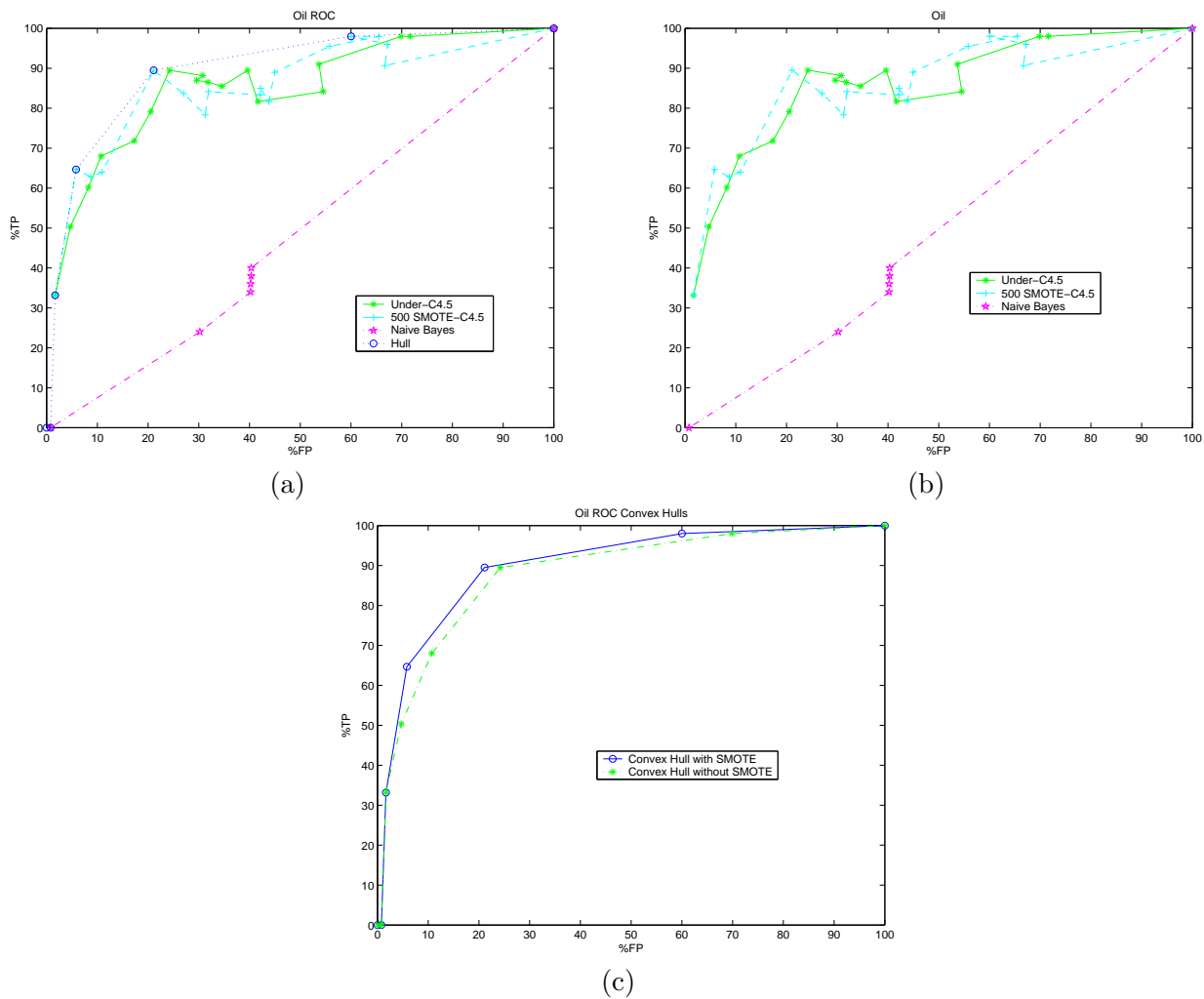

Figure 29: ROC curves for the Oil Dataset. (a) ROC curves for SMOTE-C4.5, Under-C4.5, Naive Bayes, and their ROC convex hull. (b) ROC curves for SMOTE-C4.5, Under-C4.5, and Naive Bayes. (c) ROC convex hulls with and without SMOTE.